\documentclass[10pt,twocolumn,letterpaper]{article}

\makeatletter
\def\thanks#1{\protected@xdef\@thanks{\@thanks
        \protect\footnotetext{#1}}}
\makeatother

\usepackage{iccv}
\usepackage{times}
\usepackage{epsfig}
\usepackage{graphicx}
\usepackage{amsmath}
\usepackage{amssymb}

\usepackage{amsmath}
\usepackage{algorithm}
\usepackage{algorithmic}
\usepackage{amsfonts}
\usepackage{enumerate}

\def\ie{{\em i.e.}}
\def\eg{{\em e.g.}}
\def\etal{{\em et al.}}

\newcommand{\figref}[1]{Fig. \ref{#1}}
\newcommand{\tabref}[1]{Tab. \ref{#1}}
\newcommand{\equref}[1]{(\ref{#1})}
\newcommand{\secref}[1]{Section \ref{#1}}

\newcommand{\mc}[1]{\mathcal{#1}}
\newcommand{\mb}[1]{\mathbb{#1}}

\newcommand{\tabincell}[2]{\begin{tabular}{@{}#1@{}}#2\end{tabular}}

\newcommand{\br}[1]{\bm{\mathrm{#1}}}
\newcommand{\bs}[1]{\boldsymbol{\texttt{#1}}}

\usepackage{ragged2e}
\usepackage{booktabs}
\usepackage{color}
\usepackage{multirow}
\usepackage{bm}
\usepackage{bbm}

\usepackage[pagebackref=true,breaklinks=true,letterpaper=true,colorlinks,bookmarks=false]{hyperref}

\iccvfinalcopy 

\def\iccvPaperID{1708} 

\ificcvfinal\fi

\begin{document}

\title{Heterogeneous Relational Complement for Vehicle Re-identification}

\author{  Jiajian Zhao$^{1,\dagger}$ \quad Yifan Zhao$^{1,\dagger}$   \quad  Jia Li$^{1,4,*}$ \quad Ke Yan$^{3}$   \quad Yonghong Tian$^{2,4}$\\
$^{1}$State Key Laboratory of Virtual Reality Technology and Systems, SCSE, Beihang University\\
$^{2}$Department of Computer Science and Technology, Peking University\\
$^{3}$Tencent Youtu Lab, Shanghai, China\quad $^{4}$Peng Cheng Laboratory, Shenzhen, China\\
{\tt\small \{zhaojiajian, zhaoyf, jiali\}@buaa.edu.cn, yhtian@pku.edu.cn, kerwinyan@tencent.com}
\thanks{$^{\dagger}$Jiajian Zhao and Yifan Zhao contribute equally to this work.}
\thanks{$^{*}$Jia Li is the Corresponding author. URL: \url{http://cvteam.net}}
}

\maketitle
\ificcvfinal\thispagestyle{empty}\fi

\begin{abstract}
   The crucial problem in vehicle re-identification is to find the same vehicle identity when reviewing this object from cross-view cameras, which sets a higher demand for learning viewpoint-invariant representations. In this paper, we propose to solve this problem from two aspects: constructing robust feature representations and proposing camera-sensitive evaluations. We first propose a novel Heterogeneous Relational Complement Network (HRCN) by incorporating region-specific features and cross-level features as complements for the original high-level output. Considering the distributional differences and semantic misalignment, we propose graph-based relation modules to embed these heterogeneous features into one unified high-dimensional space. On the other hand, considering the deficiencies of cross-camera evaluations in existing measures (\ie, CMC and AP), we then propose a Cross-camera Generalization Measure (CGM) to improve the evaluations by introducing position-sensitivity and cross-camera generalization penalties. We further construct a new benchmark of existing models with our proposed CGM and experimental results reveal that our proposed HRCN model achieves new state-of-the-art in VeRi-776, VehicleID, and VERI-Wild.
\end{abstract}





\section{Introduction}
Vehicle Re-identification (Re-ID) has shown broad application prospects in urban security surveillance and intelligent transportation systems. Given a gallery of images, a vehicle Re-ID algorithm aims to associate images of the same vehicle identity captured by different cameras. With the proposals of large vehicle datasets~\cite{liu2016dvehicleid,liu2016veri,lou2019veri-wild,vd,guo2018vehicle-1M} and deep learning approaches~\cite{zheng2016ide,shen2017st-path,bai2018group,tang2019pamtri,he2019part,zheng2020vehiclenet}, vehicle re-identification has made significant progresses. Although these methods achieve a performance bottleneck on existing evaluation measures, recognizing identities with large viewpoint variances, caused by the cross-camera situation, still remains a great challenge.

\begin{figure}
\begin{center}
\includegraphics[width=1\columnwidth]{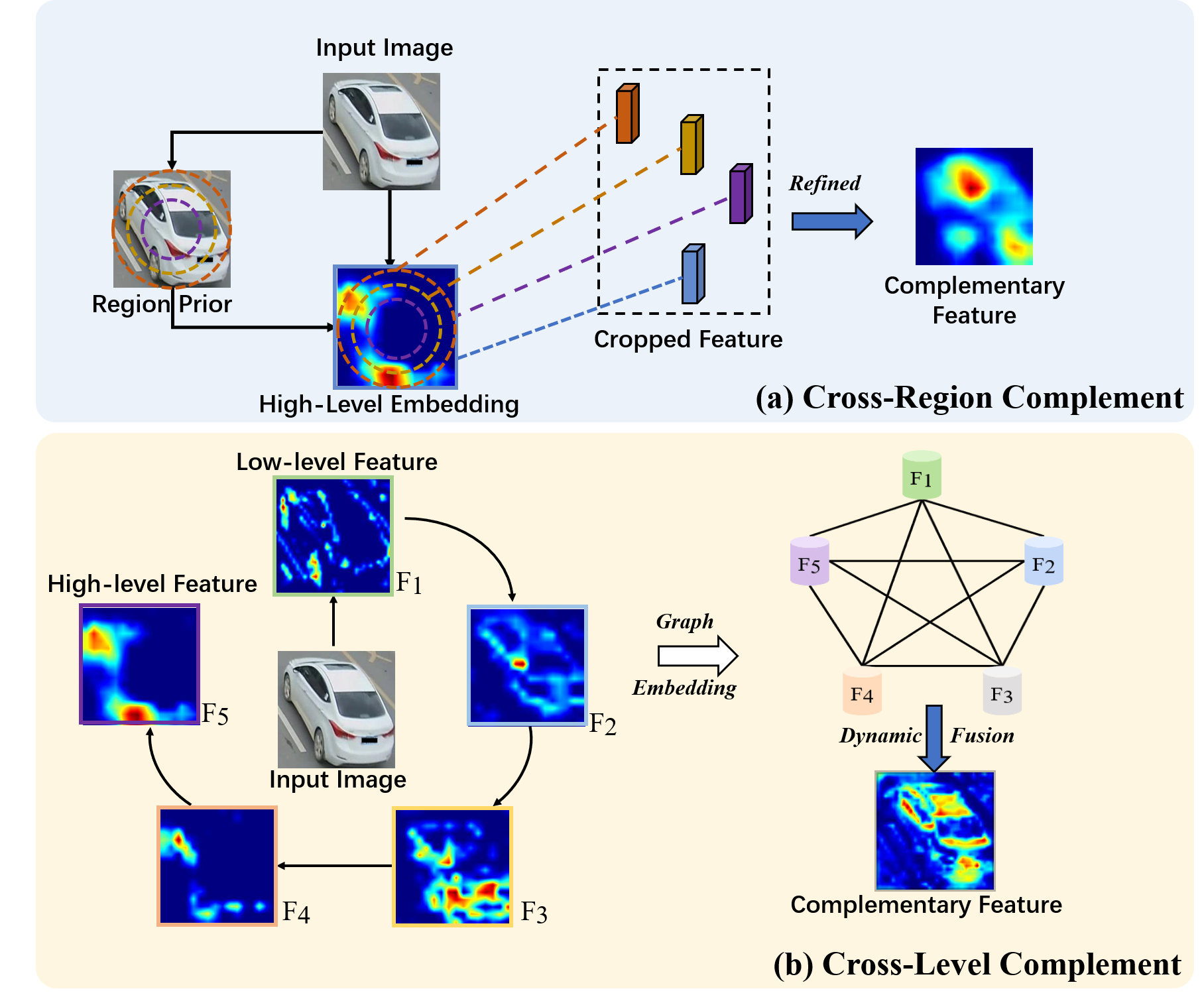}
 \caption{Motivation of our two relational complements. (a) Cross-region complement: the high-level embedding is refined by regional features to focus on discriminative regions. (b) Cross-level complement: features from multiple layers are fused by their relationship to incorporate multi-level cues.}\label{fig:motivation}
 \end{center}
 \vspace{-0.3cm}
\end{figure}

Existing researches mainly tackle this challenge from two aspects,~\ie, data-driven and feature complement. Data-driven methods deem the cross-camera challenges as a natural insufficiency of data distributions. It is observed that networks fail to recognize specific samples due to the lack of similar cases. Thus tens of recent researches tend to synthesize more examples using 3D-based models~\cite{tang2019pamtri,yao2020VehicleX} or adversarial learning~\cite{zhou2018aware,lou2019embedding,bai2020disentangled,choi2020hi-cmd}. In~\cite{yao2020VehicleX}, VehicleX dataset is composed of synthetic vehicles rendered by the Unity 3D engine. Lou~\etal~\cite{lou2019embedding} propose to generate hard negative and multi-view samples as a training data supplement. Due to their instability in synthesizing unrealistic samples, methods of this category do not explicitly regularize the feature representations for cross-camera generalization.

Methods of the second category propose to utilize discriminative regional features as a complement to the global backbone features. Recent researches for localizing these regional features usually resort to additional annotations, including keypoint localization~\cite{wang2017orientation,zhu2020structuredgraph}, bounding boxes~\cite{he2019part,zhang2020pgan} and part segmentation~\cite{meng2020parsing,liu2020beyond}. For example, He~\etal~\cite{he2019part} introduce part-regularized local features as a complement to the global representation.  Methods of this category show two major benefits: 1) strengthening discriminative regions for distinguishing subtle differences, 2) aligning parts of cross-view samples for the same identity. However, these methods heavily rely on exhaustive and accurate part annotations. Inaccurate localized part features would lead to a severe misalignment for feature embedding.


Toward these problems, in this paper, we investigate two essential cues for constructing robust complementary features. Inspired by aforementioned part-guided methods~\cite{wang2017orientation,he2019part,meng2020parsing}, we propose to learn the regional complementary embedding without any annotations, which extracted vehicle parts in a circular manner (\figref{fig:motivation} (a)). We argue that concentrating on apriori vehicle parts is beneficial for various viewpoint transformations and makes the high-level feature be aware of discriminative regions, which is desired for fine-grained identification. On the other hand, high-level features in Re-ID usually focus on limited local regions or background noises. Compared with high-level features, though low-level features lack the ability to highlight crucial regions, they contain abundant semantic information in a whole vehicle. To this end, in~\figref{fig:motivation} (b), we incorporate cross-level features from different network stages as a complement for final embeddings.


Although it seems meaningful to fuse these cross-region and cross-level features, simple aggregation strategies (\eg, concatenation and summation) on these heterogeneous features would lead to a severe misalignment, owing to the various semantics and distributions of different features. Moreover, unlike the object detection or segmentation tasks, features from different levels or even regions do not play a static role (\eg, providing clear boundaries) in recognition tasks. The same feature of different object identities shows its characteristics when constructing the final representations. Keeping this in mind, we propose a novel Heterogeneous Relation Complement Network (HRCN) to construct dynamic relations for fusing cross-level and cross-region features. To learn these two relationships, we propose a graph-based relation module to learn dynamic projections into a new feature embedding. In the cross-level complementary branch, we construct a hierarchically dynamic fusion relationship from lower to higher levels, encouraging the semantic complement for high-level ones. In the regional complementary branch, beyond the projects of part priors, we make a conjunction of the cross-level feature with the regional part features, forming a joint relation-aware representation for final classification.

Thinking from another view in the cross-camera challenge, existing measures in ReID,~\ie, CMC and AP, usually neglect the distribution of retrieved camera IDs. These measures tend to present high scores when a few samples in similar viewpoints are retrieved. To solve the natural deficiency in existing measures, we propose a new measure called Cross-camera Generalization Measure, namely CGM. This measure introduces two major factors into consideration: 1) position-sensitivity: penalizing the earlier mistakes with higher importance in each camera-independent query, 2) cross-camera generalization: treating the query on each camera as in individual retrieval task.

Contributions of this paper are three-fold:
1) We propose a novel Heterogeneous Relational Complement Network to fuse high-level features with heterogeneous complementary features, which are multi-level features and regional features, based on their relation into a robust representation feature.
2) We design a new measure named cross-camera generalization measure to evaluate more reasonably the cross-camera generalization capability of models.
3) We perform extensive experiments to reveal the proposed method outperforms state-of-the-arts on VehicleID~\cite{liu2016dvehicleid}, VeRi-776~\cite{liu2016veri} and VERI-Wild~\cite{lou2019veri-wild}, and build a benchmark of existing models with our proposed measure, CGM.

\begin{figure*}
\begin{center}
\includegraphics[width=0.9\textwidth]{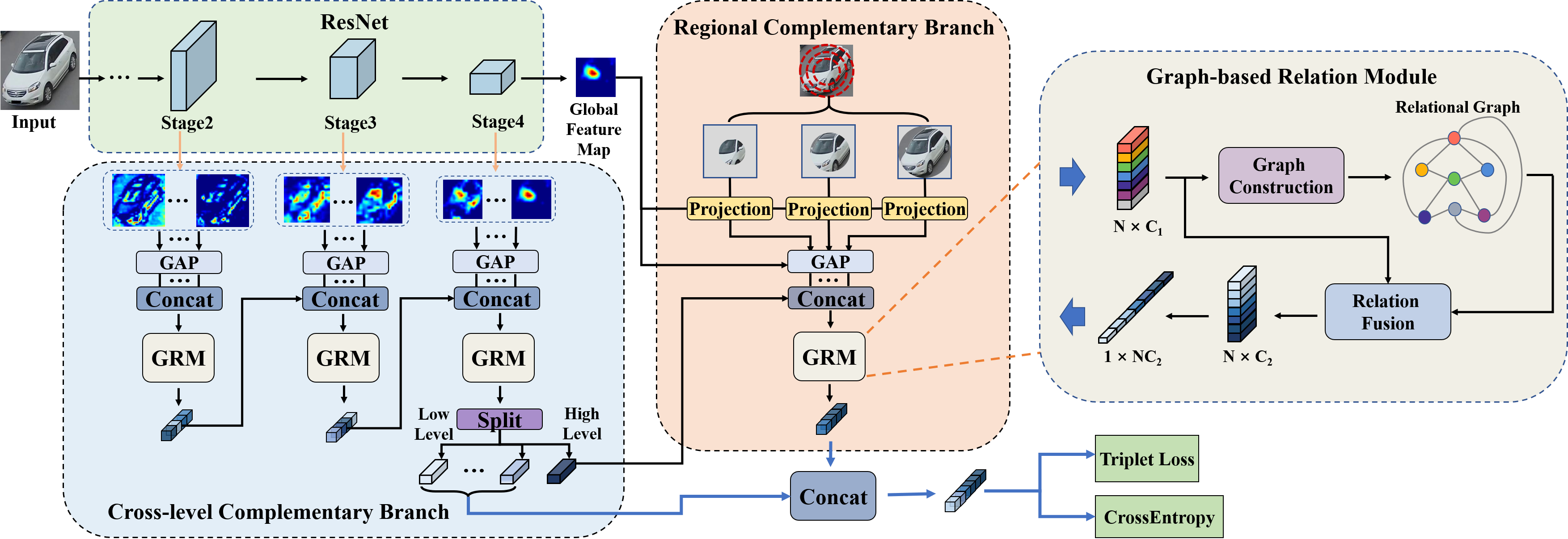}
 \caption{The pipeline of our framework. In the cross-level complementary branch, a hierarchically dynamic fusion relationship is constructed from lower to higher levels stage by stage to promote the semantic complement for high-level ones. In the regional complementary branch, beyond the projections of part priors, a conjunction of the cross-level feature with the regional part features is established to form joint relation-aware representations for final classification. To learn the dynamic relationships, graph-based relation modules are embedded into two branches to learn a dynamic projection into a new feature embedding.}\label{fig:pipeline}
 \end{center}
\end{figure*}


\section{Related Work}
\textbf{Vehicle Re-ID.} Existing vehicle Re-ID methods can be divided into four categories.
(1) Metric learning~\cite{bai2018group,Zheng2018rss,chen2019metalearning,chu2019VANet}: In~\cite{bai2018group}, GST loss was proposed to tackle the intra-class divergences by partitioning samples within each vehicle identity into a few groups. Chu~\etal~\cite{chu2019VANet} presented a viewpoint-aware metric to learn the extreme viewpoint variation issue.
(2) Multi-model fusion~\cite{shen2017st-path,wang2017orientation}: Shen~\etal\cite{shen2017st-path} introduced Siamese network with CNN and LSTM into vehicle Re-ID, which adopted an auxiliary path to explore the complex spatio-temporal regularization.
(3) Adversarial learning~\cite{zhou2018aware,lou2019embedding,bai2020disentangled}: In~\cite{zhou2018aware}, a viewpoint-aware attentive generation model was proposed to generate vehicles of specific angles to help understand the cross-view features.
(4) Part complement~\cite{he2019part,meng2020parsing,liu2020beyond}: In~\cite{wang2017orientation}, vehicle landmarks were used to extract local orientation-invariant features. He~\etal~\cite{he2019part} used a detection model to find the bounding boxes of some parts in a vehicle, which was beneficial to tackle the near-duplicate phenomenon in vehicle re-identification. In~\cite{meng2020parsing,liu2020beyond}, to locate the precise local parts of vehicles,
a semantic segmentation network was proposed to obtain pixel-level part localization. Although these methods achieved better performance, it requires huge labor costs to label additional annotations. In this paper, we propose cross-level and region-specific features as complements instead of annotated part features.

\textbf{Vehicle Re-ID Measures.} Existing measures~\cite{zheng2015scalable,liu2016dvehicleid} in Re-ID are based on query results of a ranking list. Zheng~\etal~\cite{zheng2015scalable} proposed average precision (AP) in the Re-ID task to combine two classical classification measures, \ie, precision and recall. Liu~\etal~\cite{liu2016dvehicleid} introduced cumulative matching characteristics (CMC) into the vehicle Re-ID.
To eliminate the impact of images captured from the same camera, Liu~\etal~\cite{liu2016veri} proposed an image-to-track metric named HIT. However, these aforementioned measures ignore the distribution of retrieved camera IDs.


\textbf{Graph Convolution Network.} Graph convolution network has broadened its applications in computer vision tasks~\cite{wang2019faceGCN,hu2020segGCN}. Kipf~\etal~\cite{kipf2016GCN} proposed a simple but effective spatial graph convolution, which was a localized first-order approximation of spectral graph convolutions. In~\cite{velivckovic2017GAT}, a self-attention mechanism was introduced into the graph neural network to overcome the shortcoming that GCN depends on the fixed graph structure. Li~\etal~\cite{li2019deepgcns} utilized a residual connection to deal with the gradient vanishing in deep GCNs.  For video-based re-identification, Yang~\etal~\cite{yang2020st-graph} proposed a spatial-temporal graph convolutional network to extract and combine the spatial-temporal information to tackle problems of occlusion and ambiguity. In this paper, we propose a new graph-based relation module to embed heterogeneous features into a unified high-dimensional space instructed by the relation.
\section{Approach}
\subsection{Overview}



In this section, we introduce a Heterogeneous Relation Complement Network (HRCN) to construct dynamic graph-based relationship for regional and cross-level feature complements. As in~\figref{fig:pipeline}, our key idea is to build dynamic learnable relationships for these region-specific and layer-specific features. Given an input image $\mc{I}$, let $\br{V}^{i} \in \mathbb{R}^{N \times C}$ be the concatenated feature of $N$ layers in the $i$th network stage, which is squeezed by the global pooling operation. We first build a hierarchical cross-level relationship in~\secref{sect:cross-level branch}, by using the proposed graph-based relation module $\mc{G}$ in~\secref{sect:relation module}. We then extract features from the second stage, forming a cross-layer fusion of $\mc{G}^{2}(\br{V}^{2})$. Following this manner, features from higher stages can also be exploited by building $\mc{G}^{3}(\mc{G}^{2}(\br{V}^{2});\br{V}^{3})$ and so on. We denote the $n$th dynamic aggregation of the feature set $\mathbb{V}$ as $\mc{G}^{(n)}(\mathbb{V})$. Beyond the cross-level relationship, we further propose a regional feature complement in~\secref{sect:regional branch} for enhancing $S$ prior regional features $\br{f}^{r}$ without additional annotations. Thus the heterogeneous features from these two branches are fused to achieve the final embedding $\br{E}$ with the relational conjunction $\mc{R}(\cdot)$:
\begin{equation}\label{overall}
\br{E}=\mc{R}(\mc{G}^{(n)}(\mathbb{V});\{ \br{f}^{r}_{k}  \}_{k=1}^{S}).
\end{equation}
Hence a dynamic fusion process of heterogeneous features can be established by exploiting the learnable graph relations. It endows the network a self-directed capability to dynamically select the informative features based on the semantic relationships.


\subsection{Cross-level Feature Complement}\label{sect:cross-level branch}
Despite the lack of the ability to highlight specific regions, low-level features contain abundant semantic information of a whole vehicle due to less information loss by the operation of convolution kernels. Complemented by lower levels, high-level features can improve in two aspects: 1) paying attention to more discriminative regions. 2) avoiding falling completely into non-critical regions.

Nevertheless, under the heterogeneity of high-level features and low-level features, there exist distributional differences and semantic misalignment between them. To directly adopt common aggregations like concatenation and summation, it may lead to semantic confusion rather than achieve a positive complement. To solve the aggregation problem, we propose to project cross-level features into a unified space guided by their closeness of the relation. Setting a higher fusion coefficient to the closer one and aggregating all different levels based on the respective coefficient, the new embedding not only retains its original semantic information but also eliminates the differences among them. Considering the complementary positions of low levels, we take all features in the lower stage as a feature to the higher stage, which ensures the dominance of high levels.


Here we select the final output feature in a block as the corresponding level feature. To reduce computation and memory cost, the feature from the $j${th} block in the $i$th stage is squeezed into a vector $\br{v}_{j}^{i}$ by the global average pooling and then aligns its channels with other levels by a linear transformation. These cross-level aligned vectors will be concatenated as a whole, which is defined as
\begin{equation}\label{eq:level fusion 1}
   \mathbf{V}^{i}= \begin{cases}\mathcal{C}\left(\mathbf{W}_{1}^{i} \mathbf{v}_{1}^{i}, \ldots, \mathbf{W}_{k}^{i} \mathbf{v}_{k}^{i}\right) & i=2 \\ \mathcal{C}\left(\mathbf{W}_{1}^{i} \mathbf{v}_{1}^{i}, \ldots, \mathbf{W}_{k}^{i} \mathbf{v}_{k}^{i}, \mathbf{W}_{k+1}^{i} \mathcal{G}^{i-1}\left(\mathbf{V}^{i-1}\right)\right) & i>2\end{cases}
\end{equation}
where $k$ is the number of selected blocks in the ${i}$th stage, $\br{W}$ is a learnable weight matrix and $\mc{C}$ represents a concatenation operation.

Then the concatenated vector will be sent into a graph-based relation module to conduct a relational aggregation, which can embed heterogeneous features into the same representation space. After the fusion, the fused vector $\br{\hat{v}}^{i}$ will be taken as a low-level complementary feature into the next stage. In the last relational fusion stage, we split the final fused vector $\br{\hat{v}}^{n}$ into several vectors representing the respective level feature. The highest level vector $\br{\hat{v}}^{n}_{high}$ is sent into regional feature complementary branch to learn the complementary information from regional features, and other split vectors will be projected into different sub-space by using a $1 \times 1$ convolution according to their importance.

\subsection{Regional Feature Complement} \label{sect:regional branch}
Unlike previous works using additional hand-crafted annotations~\cite{he2019part,meng2020parsing,liu2020beyond}, we introduce a novel progressive center pooling to align different regions of multiple identities. With the locally aligned regions, models have the potential to measure the cross-view features in one unified embedding. Moreover, constructing region information using detection or segmentation models would cause instability for semantic consistency,~\ie, detectors usually fail to localize the correct parts in complex scenarios, which would lead to catastrophic over-fittings.

Remarkably, there is a meaningful prior in vehicle Re-ID that all images are strictly cropped and aligned to form holistic objects. Starting from this prior, it can be observed that the key regions usually exist in the center of one image. With the expansions of receptive field in~\figref{fig:pipeline}, richer features of one vehicle are gradually obtained but introduce more background confusions. We thus construct pyramid-centered features with $S$ local regions and then adopt ROI projection operation~\cite{he2019part} to extract its region feature without any additional computation costs.

The center pooling operation embraces discriminative regions into local features in a step-wise manner. ~\eg, the vehicle in regional complementary branch~\figref{fig:pipeline} focuses on window regions at the first pooling pyramid, and at the second pyramid, attention regions are extended to incorporate lights and the side. Under center pooling, it is observed that discriminative information contained by local features is from less to more and from centralization to generalization, making local features have generalized information.

Assuming the left bottom image corner as the origin of coordinates with a given image $\mc{I} \in \mathbb{R}^{W\times H}$, the circular center mask region $\br{M}$ of $k$th region can be formulated as:
\begin{equation}   \label{center mask}
   \mathbf{M}_{x, y}^{k}= 
   \begin{cases}1 & \text { if }\left(x-\frac{W}{2}\right)^{2}+\left(y-\frac{H}{2}\right)^{2} \leq R_{k}^{2} \\ 0 & \text { otherwise }
   \end{cases}
\end{equation}
where $R_k$ denotes the radius on $k$th circle.
With the extracted mask regions, we perform a re-projection $\mc{P}(\cdot)$ on global feature $\br{F}^{g}$ to form the regional embedding $\br{f}^{r}$:
\begin{equation}  \label{center feature}
   \br{f}^{r}_{k}= \br{W}_k \cdot \phi(\mc{P} (\br{F}^{g}, \br{M}^{k}))+\br{B}_k, k= 1\ldots S,
\end{equation}
where $\phi$ denotes the global average pooling. $\br{W}_k$ and $\br{B}_k$ denotes the learnable weights of linear transformations.

Then, these region-specific vectors and the global vector $\br{f}^{g}$ produced by squeezing $\br{F}^{g}$ are taken as heterogeneous complements for the highest level fused vector $\br{\hat{v}}^{n}_{high}$ from the cross-level complementary branch. By the graph-based relation module, a conjunction of the cross-level features with region-specific features is constructed. Taking other lower levels, a joint relation-aware embedding is formed:
\begin{equation}\label{final embedding}
\br{E}=\mc{C}(\{ \br{\hat{v}}^{n}_{l} \}_{l=1}^{high-1},\mc{G}(\mc{C}(\br{\hat{v}}^{n}_{high},\br{f}^{g},\{ \br{f}^{r}_{k} \}_{k=1}^{S}))),
\end{equation}
where $\br{\hat{v}}^{n}_{l}$ is the ${l}$th level fused vector from the $n$th stage.

\subsection{Graph-based Relation Module}\label{sect:relation module}
\begin{figure}
\begin{center}
\includegraphics[width=1\columnwidth]{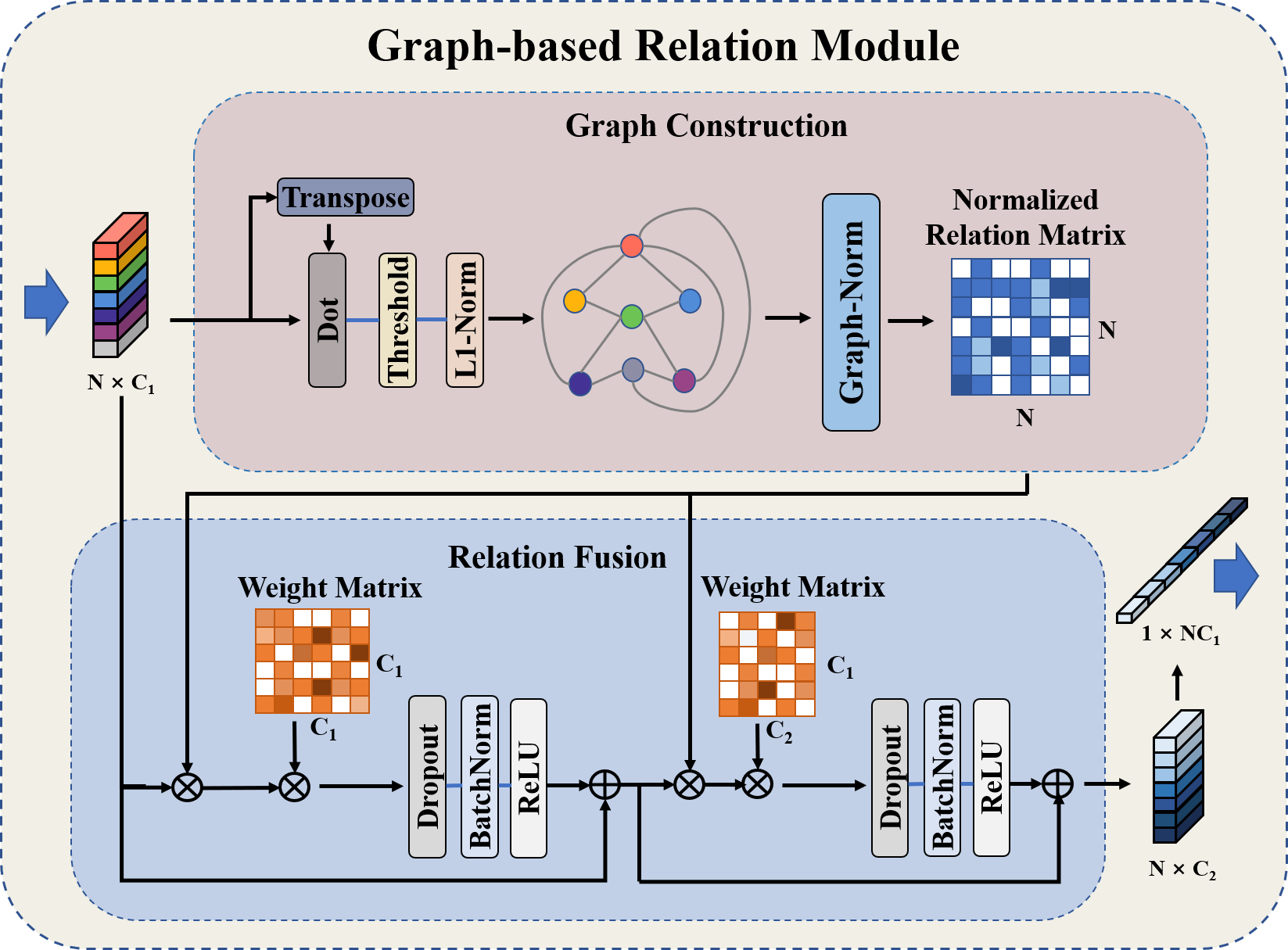}
 \caption{Illustration of graph-based relation module. The proposed module consists of two parts which are graph construction and relation fusion. In graph construction, we build a normalized relation matrix among different features. In the relation fusion, we set two weight matrices with different sizes. One is utilized to keep channel-invariant relational fusion and the other one aims to realize a channel-squeezed relational fusion.}\label{fig:module}
 \end{center}
\end{figure}

In this section, we introduce the construction process of graph-based relation module consisting of the graph construction and the relation fusion, which is shown in~\figref{fig:module}. The crucial problem in graph construction is how to compute the dynamic relational edge and determine the edge connection style. For constructing this edge, we take the similarity between two features as their relational weight, which is computed as:
\begin{equation}  \label{kernel}
    \br{A}(\br{v}_{i},\br{v}_{j})=\br{v}_{i}\br{v}_{j}^{T},
\end{equation}
where $\br{A} \in \mb{R}^{N \times N}$, $\br{A}(\br{v}_{i},\br{v}_{j})$ represent the edge weight between the $i$th feature and the $j$th feature.

After the edge weight is computed, a complete graph will be constructed. However, a complete relational graph not only requires a high computational cost, but also degrades the representation ability of the graph due to losing the transitional effect for intermediate nodes. To sparse the complete graph, we eliminate some relational edges by setting a threshold, which can be formulated as:
\begin{equation}   \label{threshold}
\mathbf{A}^{\prime}\left(\mathbf{v}_{i}, \mathbf{v}_{j}\right)= \begin{cases}\mathbf{A}\left(\mathbf{v}_{i}, \mathbf{v}_{j}\right) & \text { if } \mathbf{A}\left(\mathbf{v}_{i}, \mathbf{v}_{j}\right) \geq \alpha \\ 0 & \text { otherwise }\end{cases}
\end{equation}
where $\alpha$ is a hyperparameter standing for the threshold.

Then we adopt L1-normalization to restrain the edge weight in each row within the range of $(0,1)$. Avoiding losing the original semantic information, we add an identity matrix $\br{I}_{n} \in \mb{R}^{N \times N}$ to the relational matrix $\br{A}^{'}$ to keep the dominance in itself, and then a re-normalization trick~\cite{kipf2016GCN} is utilized to approximate the graph-Laplacian:
\begin{equation}  \label{graph_norm}
    \br{\hat{A}}=\br{D}^{-\frac{1}{2}}(\br{A'}+\br{I}_{n})\br{D}^{-\frac{1}{2}},
\end{equation}
where $\br{\hat{A}}$ is the normalization relation matrix, $\br{D}$ is a diagonal matrix and $\br{D}_{i,i}=\sum_{j}^{N} \br{A'}_{i,j} + \br{I}_{i,i}$.

For the relation fusion in~\figref{fig:module}, we propose a two-step fusion relied on the normalized relation matrix $\br{\hat{A}}$ and learnable weight matrices. In the first step, heterogeneous features will multiply by the relation matrix to eliminate discrepancies among features in a unified high-dimensional space. Subsequently, a weight square matrix is set to be multiplied by relational features, to keep a channel-invariant characteristic that avoids losing more information in the training. In the second step, we not only promote the second relational fusion, but also squeeze dimensions of features by adjusting the size of the weight matrix. To prevent overfitting, a dropout layer is added after the weight matrix multiplication. Meanwhile, we take a residual connection between the input and the output in each step to restrain the vanishing gradient. The fusion process for output $\br{O}$ in each step can be defined as:
\begin{equation}
    \br{O}=\bs{ReLU}(\bs{BN}(\bs{Dropout}(\br{\hat{A}}\br{V}\br{W}_{r}))) + \br{W}_{a}\br{V},
  \label{graph_norm}
\end{equation}
where $\br{W}_{r}$ is the weight matrix for channel transformation and $\br{W}_{a}$ is a learnable tensor, aligning feature channels in $\br{V}$ with the output $\br{O}$.


\section{Measure}

\subsection{Limitations of Current Measures}

\textbf{Cumulative Matching Characteristics.} The CMC curve indicates the probability that the query identity appears in the different-sized interval of the retrieve list. CMC curve can be formally represented as $CMC@k=\frac{\sum_{i=1}^{N}m(q_{i},k)}{N}$. $N$ denotes the number of queries and $\mathit{q_{i}}$ is the identity of the $\mathit{i^{th}}$ query. $\mathit{m(q_{i},k)}$ is $1$ if the target identity appears in the interval of the rank list from top-$1$ to top-$k$, otherwise it is equal to $0$.
CMC is a binary measure for its focusing on whether the query identity exists in the specific interval, which cannot evaluate the distribution of the target images in the holistic rank list. Despite the subtle performances of multiple methods, CMC is only suitable for the gallery where each identity has very few samples, yet it cannot evaluate the cross-camera generation at all.

\textbf{Average Precision.} AP is proposed to measure the distribution of a ranking list. It provides a joint description of precision and recall for the classification results. Therefore, the calculation of AP is based on the integral of the Precision-Recall (P-R) curve, which can be defined as $AP(q)=\sum_{k=1}^{N}P(k)\Delta r(k)$. $\mathit{P(k)}$ is the precision at a cutoff of $\mathit{k}$ images. $\mathit{k}$ denotes the total number of gallery images, and $\mathit{\Delta r(k)}$ is the difference in recall that occurs between cutoff $\mathit{k}$ and $\mathit{k-1}$.
Based on the precision and recall, AP encourages correct matched pairs and penalizes the error pairs, which equally treats the precision of each target image in the list. Therefore, AP constructs a global scope of the distribution of all target images in the retrieval list.

Here we explore two deficiencies of AP measure for Re-ID tasks:
1) AP neglects the consideration of different cameras where the target images are captured, due to measuring all target images as a whole list.
2) AP measure is not sensitive to where the errors occur in the retrieval list. AP shows subtle value changes regardless of where the error position appears in the ranking list.

\begin{figure}
\begin{center}
\includegraphics[width=1\columnwidth]{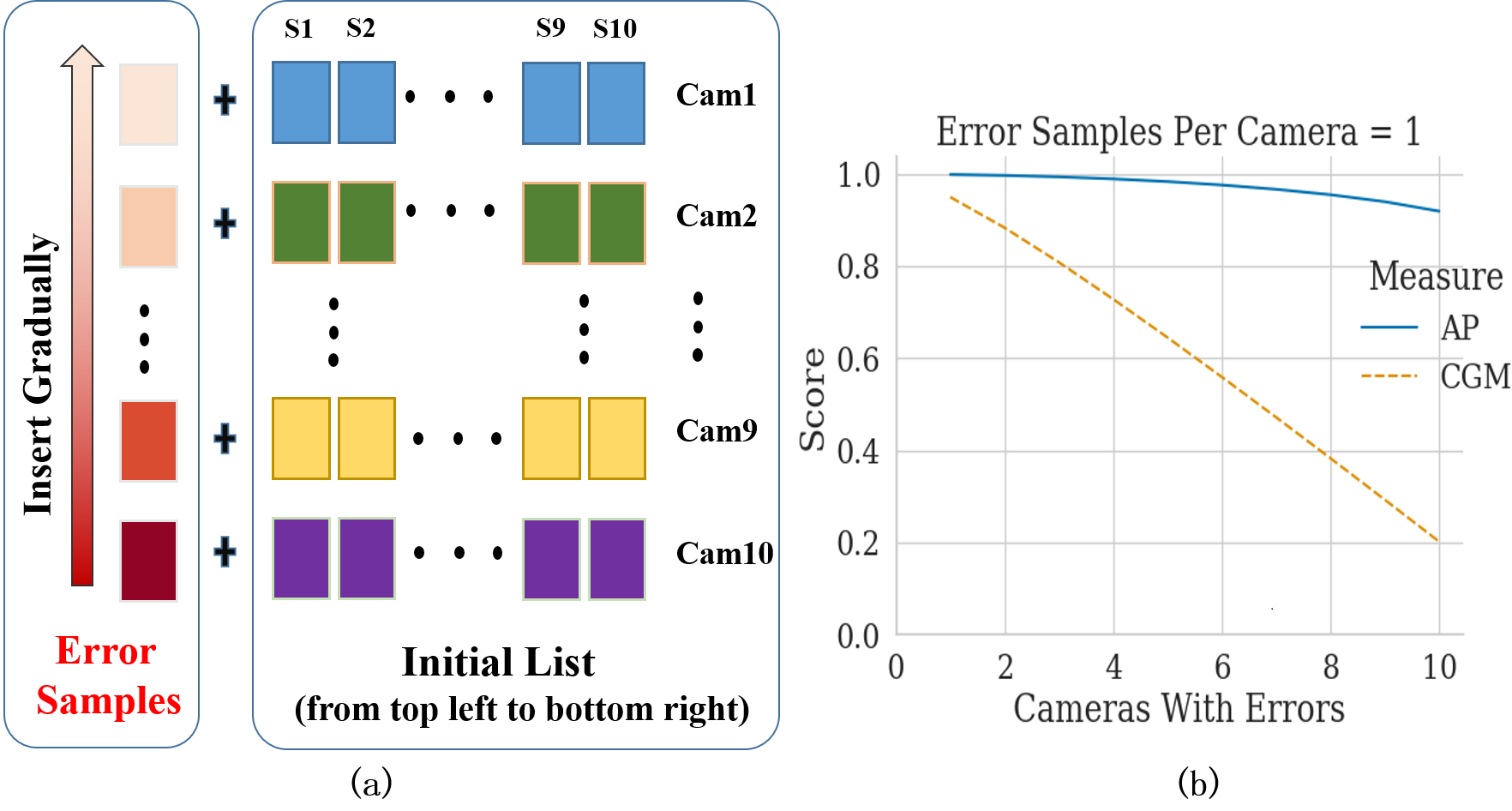}
 \caption{Cross-camera generalization capability comparison between AP and CGM. Here we assume an initial list of 100 correct samples averagely captured from 10 cameras. (a) Illustration of our setting: given these 10-camera sub-galleries, we sequentially insert 1 error sample into each individual camera sub-gallery from back to front. (b) With the increase of error samples, the scores on backward camera sub-gallery should decrease sharply, indicating the deterioration of cross-camera generalization. AP performs subtle changes while proposed CGM declines evidently. } \label{fig:AP_CGM_Cross_Cams}
 \end{center}
\end{figure}

\subsection{Cross-camera Generalization Measure}

~\textbf{Definition.}  A problem aforementioned measures are the lack of consideration on the cross-camera generalization. To this end, a new measure should make sure every camera performs independent effects on the final score. Hence the proposed CGM first divides target images captured from the same cameras into individual groups.~\ie, CGM independently measures the ranking result of each camera by removing target images captured from other cameras from the ranking list, forming an individual sub-gallery. Given one query identity $\mathit{q}$ and one specific camera $\mathit{C_{i}}$, the proposed $1$-query $1$-camera CGM has the form:
\begin{equation}  \label{CGM@C}
  CGM(q,C_{i})=\frac{1}{{N_{C_{i}}}}\sum_{k=1}^{N_{C_{i}}}\frac{1}{E(k)+1},
\end{equation}
where $\mathit{N_{C_{i}}}$ denotes the number of target images captured from the $\mathit{C_{i}}$ and $\mathit{E(k)}$ denotes the summation of error samples before the $\mathit{k}$th target image in the ranking list.

Considering the performance of all cameras, the $1$-query $N_C$-camera CGM can be defined as:
\begin{equation}  \label{CGM}
  CGM(q)=\frac{\sum_{i=1}^{N_{C}}CGM(q,C_{i})}{{N_{C}}}.
\end{equation}
For a generalized Re-ID task with $N$-query $N_C$-camera setting, we propose the mean Cross-camera Generalization Measure (mCGM):
\begin{equation}  \label{mCGM}
  mCGM=\frac{\sum_{q=1}^{N}CGM(q)}{N}.
\end{equation}
With this definition, here we elaborate on two insights to reveal the findings on measuring vehicle Re-ID tasks.

\textbf{Position-sensitive Capability.}
As mentioned above, the earlier mistake in the retrieval list should be attached with more importance. Each error sample should leave a negative impact on all samples after its position. Nevertheless, AP focuses on the correctness of the whole gallery and is thus insensitive to the error position. To solve this issue, CGM adopts a discounted strategy that takes the summation of error samples ahead of the target sample as a penalty factor (denoted as $\frac{1}{E(k)+1}$ in Eq.~\equref{CGM@C}). Hence two advantages are introduced by adopting this discounted strategy: 1) It can linearly transmit the penalty to correct samples after the error position, which avoids the decay of the impact of forwarding error positions to backward positions. 2) The gradient of the penalty factor computed by $\frac{1}{E(k)+1}$ decreases gradually with the increase of error samples. This improves the impact of forwarding errors that makes the measure more sensitive to error positions.

\textbf{Cross-camera Generalization Capability.} To consider the cross-camera generalization, CGM calculates the score on each camera independently by removing target images captured from the other cameras in the final ranking list in Eq.~\equref{CGM@C}, and averages values of each camera in Eq.~\equref{CGM}. In~\figref{fig:AP_CGM_Cross_Cams}, we illustrated that CGM declines sharply when the errors are gradually inserted into the correct list, while AP performs in a non-obvious manner.

Compared with AP, the proposed CGM not only focuses on the holistic ranking distributions, but also on the sensitivity of error positions and performance independently on each camera. The positional sensitivity makes the evaluation score determined by the correct ranking rather than the correct proportion of the whole list. The dependence of each camera ensures CGM a camera-level measure that solidly meets the demands of cross-camera generalization.

\begin{table}[t]
\centering{
\caption{Performance comparisons on VeRi-776 benchmark. }
\label{table:performance on VeRi}
\renewcommand{\arraystretch}{0.8}
\begin{tabular}{l|c|c|c}
\toprule
Method&CMC@1&CMC@5&mAP\\
\midrule
BOW-CN~\cite{zheng2015scalable}&0.339&0.537&0.122\\
LOMO~\cite{liao2015person}&0.253&0.465&0.096\\
FACT~\cite{liu2016large}&0.510&0.735&0.185\\
GoogLeNet~\cite{yang2015large}&0.498&0.712&0.170\\
RAM~\cite{liu2018ram}&0.886&0.940&0.615\\
EALN~\cite{lou2019embedding}&0.844&0.941&0.574\\
VAMI~\cite{zhou2018aware}&0.770&0.908&0.501\\
OIFE~\cite{wang2017orientation}&0.894&-&0.480\\
AAVER~\cite{khorramshahi2019aaver}&0.890&0.947&0.612\\
PRN~\cite{he2019part}&0.943&0.987&0.743\\
PVEN~\cite{meng2020parsing}&0.956&0.984&0.795\\
\cline{1-4}
\midrule
Baseline&0.954&0.980&0.768\\
Only Regional branch &0.966&0.986&0.807\\
Only Cross-level branch&0.964&0.988&0.814\\
Ours (w/o GRM)&0.965&0.984&0.803\\
Ours (full)&\textbf{0.973}&\textbf{0.989}&\textbf{0.831}\\
\bottomrule
\end{tabular}
}
\end{table}

\begin{table}
\centering{
\caption{Performance comparisons on VehicleID.}
\label{table:CMC on VehicleID}
\resizebox{\columnwidth}{!}{
\begin{tabular}{l|c|c|c|c|c|c}
\toprule
\multirow{2}*{\tabincell{c}{Method}}& \multicolumn{2}{|c|}{Small}& \multicolumn{2}{|c|}{Medium}& \multicolumn{2}{|c}{Large}\\
\cline{2-7}
&C@1&C@5&C@1&C@5&C@1&C@5\\
\midrule
DRDL~\cite{liu2016dvehicleid}&0.490&0.735&0.428&0.668&0.382&0.616\\
RAM~\cite{liu2018ram}&0.752&0.915&0.723&0.870&0.677&0.845\\
EALN~\cite{lou2019embedding}&0.751&0.881&0.718&0.839&0.693&0.814\\
VAMI~\cite{zhou2018aware}&0.631&0.833&0.529&0.751&0.473&0.703\\
AAVER~\cite{khorramshahi2019aaver}&0.747&0.938&0.686&0.900&0.635&0.856\\
OIFE~\cite{wang2017orientation}&-&-&-&-&0.670&0.829\\
PRN~\cite{he2019part}&0.784&0.923&0.750&0.883&0.742&0.864\\
PVEN~\cite{meng2020parsing}&0.847&0.970&0.806&0.945&0.778&0.920\\
\cline{1-7}
Baseline&0.762&0.894&0.763&0.874&0.731&0.848\\
Ours&\textbf{0.882}&\textbf{0.984}&\textbf{0.814}&\textbf{0.966}&\textbf{0.802}&\textbf{0.944}\\
\bottomrule
\end{tabular}
}
}
\end{table}

\begin{table}
\centering{
\caption{Performance comparisons on VERI-Wild. $\ast$: not removing gallery images of the same identity and camera as querying.}
\label{table:perfomance on VERI-Wild}
\setlength{\tabcolsep}{0.4mm}
\renewcommand{\arraystretch}{1.15}
\resizebox{\columnwidth}{!}{
\begin{tabular}{l|c|c|c|c|c|c|c|c|c}
\toprule
\multirow{2}*{\tabincell{c}{Method}}& \multicolumn{3}{|c|}{Small}& \multicolumn{3}{|c|}{Medium}& \multicolumn{3}{|c}{Large}\\
\cline{2-10}
&C@1&mAP&mCGM&C@1&mAP&mCGM&C@1&mAP&mCGM\\
\midrule
FDA~\cite{lou2019veri-wild}&0.640&0.351&-&0.578&0.298&-&0.494&0.228&-\\
AAVER~\cite{khorramshahi2019aaver}&0.758&0.622&-&0.682&0.537&-&0.587&0.417&-\\
BW~\cite{kuma2019tribaseline}&0.842&0.705&-&0.782&0.628&-&0.700&0.516&-\\
SAVER~\cite{khorramshahi2020saver}&0.945&0.809&-&0.927&0.753&-&0.895&0.677&-\\
PGAN~\cite{zhang2020pgan}&0.951&0.836&-&0.928&0.783&-&0.892&0.706&-\\
PVEN~\cite{meng2020parsing}$^{\ast}$&\textbf{0.967}&0.825&-&\textbf{0.954}&0.770&-&\textbf{0.934}&0.697&-\\
\midrule
Ours&0.940&\textbf{0.852}&\textbf{0.759}&0.916&\textbf{0.800}&\textbf{0.692}&0.880&\textbf{0.722}&\textbf{0.601}\\
\bottomrule
\end{tabular}
}
}
\end{table}

\section{Experiments}
\subsection{Experimental Settings}
\textbf{Datasets.}
1) VeRi-776~\cite{liu2016veri} is the most widely-used vehicle Re-ID benchmark, which has over 50,000 images of 776 vehicles. Vehicles in VeRi-776 are captured from 20 cameras in multiple viewpoints, hence we use this dataset to measure the cross-camera generalization.
2) VehicleID~\cite{liu2016dvehicleid} consists of 211,763 images with 26,267 vehicles under the front or back viewpoint. In evaluation, three test subsets (\ie, small, medium and large) are split according to their sizes. For each subset, we randomly select one image of each identity into the gallery set and the rest images are taken as query set.
3) VERI-Wild~\cite{lou2019veri-wild} is large-scale dataset containing 416,314 images of 40,671 vehicles from 174 cameras. Images in VERI-Wild involve various viewpoints, weathers and illuminations. There are 138,517 images of 10,000 identities in the test set, which consists of three subsets with 3,000, 5,000, 10,000 identities respectively.

\textbf{Training details.}
We use ResNet-50~\cite{resnet} with ibn-a blocks~\cite{pan2018ibn} pretrained on ImageNet as our backbone. The model is trained for 120 epochs with the SGD optimizer. We warm up the learning rate from 7.7e-5 to 1e-2 in the first 20 epochs and the backbone is frozen in the warm-up step. The learning rate of 1e-2 is kept until the 60th epoch and drops to 7.7e-5 under a cosine annealing schedule in the rest epochs. The batch size is 48 (12 IDs, 4 instances) in VehicleID and VeRi-776, and 120 (30 IDs, 4 instances) in VERI-Wild. We pad 10 pixels on the image border, and then randomly crop it to 256$\times$256. Random erasing is taken to augment the data and the threshold $\alpha$ is set as 1e-3.

\begin{table}[t]
\centering{
\caption{A benchmark of 11 models with 3 measures on VeRi-776 dataset. Ex Ans: extra annotations.}
\label{table:Benchmark}
\setlength{\tabcolsep}{1.0mm}
\begin{tabular}{l|c|c|c|c}
\toprule
Method&Ex Ans&mCGM&mAP&CMC@5 \\
\midrule
S-Baseline~\cite{luo2019strong}&&0.511&0.744&0.978\\
+Triplet Loss&&0.543&0.764&0.978\\
+CCL&&0.547&0.769&0.975\\
IDE~\cite{zheng2016ide}&&0.415&0.665&0.975\\
DMML~\cite{chen2019metalearning}&&0.501&0.697&0.955\\
VDK~\cite{porrello2020robust}&\checkmark&0.543&0.775&0.981\\
PAMTRI~\cite{tang2019pamtri}&\checkmark&0.503&0.684&0.968\\
RECT-Net~\cite{zhu2020voc}&&0.562&0.778&0.968\\
PVEN~\cite{meng2020parsing}&\checkmark&0.560&0.783&0.983\\
\cline{1-5}
Baseline&&0.554&0.768&0.980\\
Ours&&\textbf{0.630}&\textbf{0.831}&\textbf{0.989}\\
\bottomrule
\end{tabular}
}
\end{table}

\subsection{Comparisons with The State-of-the-arts}
~\textbf{Compared Methods.} BOW-CN~\cite{zheng2015scalable} and LOMO~\cite{liao2015person} are handcrafted feature-based methods. FACT~\cite{liu2016large} takes fusion between the handcrafted and deep CNN features. DRDL~\cite{liu2016dvehicleid}, BW~\cite{kuma2019tribaseline} and RAM~\cite{liu2018ram} adopt a deep CNN to learn visual features from holistic appearances. In~\cite{lou2019embedding,zhou2018aware,lou2019veri-wild}, adversarial scheme is used to gain a robust feature. SAVER~\cite{khorramshahi2020saver} extracts a vehicle-specific feature aided by VAE~\cite{kingma2013vae}. AAVER~\cite{khorramshahi2019aaver}, OIFE~\cite{wang2017orientation}, PRN~\cite{he2019part}, PGAN~\cite{zhang2020pgan} and PVEN~\cite{meng2020parsing} utilize extra annotations to train a specific network to locate regions of interest, which include discriminative features of a vehicle.

~\textbf{VeRi-776 Benchmark.} To evaluate the effectiveness, we adopt three measures, which are CMC@1, CMC@5 and mAP, to compare our model with 11 state-of-the-art methods. Relied on extra annotations, these methods~\cite{he2019part, meng2020parsing} achieve a large promotion on VeRi-776. However, limited by the quality and the number of annotations, the performance of the above methods is hard to be further improved. Our proposed method adopts heterogeneous complementary features to aid the network to learn a robust representation. As shown in~\tabref{table:performance on VeRi}, our proposed approach without extra annotations outperforms all methods even if there is only a cross-level complementary branch or regional complementary branch in the model. Compared with only one branch, the conjunction of two branches can improve the performance appreciably, especially in mAP.

~\textbf{VehicleID Benchmark and VERI-Wild Benchmark.}  As shown in~\tabref{table:CMC on VehicleID}, we compare our models and 8 state-of-the-art methods with two criteria,~\ie, {CMC@1 and CMC@5}. Without any extra annotations, our model outperforms PVEN~\cite{meng2020parsing} at CMC@1 and CMC@5 on three test subsets. It demonstrates our approach achieves excellent performance in the extreme viewpoint environment. On VERI-Wild, our proposed method is compared with 6 state-of-the-arts in CMC@1 and mAP. In~\tabref{table:perfomance on VERI-Wild}, it is observed that our approach surpasses all methods at mAP, which displays that HRCN has a robust generalization ability in the large-scale dataset. Meanwhile, we exhibit mCGM of our method in three test subsets on VERI-Wild.

\subsection{A Benchmark Analysis of Joint Measures}
To demonstrate our proposed measure, we construct a benchmark with CMC, mAP and mCGM in~\tabref{table:Benchmark}. We collect open-source models and follow their official training strategy to train them from scratch under the training dataset in VeRi-776.
S-Baseline~\cite{luo2019strong} is a ResNet-50 backbone with a batch normalization layer before softmax loss. IDE is the ID-discriminative embedding proposed by~\cite{zheng2015scalable}, and it is implemented by ~\cite{yao2020VehicleX} in our proposed benchmark. DMML~\cite{chen2019metalearning} is a meta metric learning which divides samples with the same identity into several meta groups. VDK~\cite{porrello2020robust} adopts a self-distillation by transferring a video-based network to an image-based network. In~\cite{tang2019pamtri}, we take its multi-learning network trained by labels with identity, color, and type. RECT-Net~\cite{zhu2020voc} stands for ResNet backbone, Generalized-mean Pooling, Circle loss, and Triplet loss. In~\cite{meng2020parsing}, extra annotations are adopted to train a specific network to locate more precisely the regions of interest.

According to~\tabref{table:Benchmark}, it is obvious that our proposed mCGM is not positive-correlated with mAP and is a harder measure than mAP. We can also observe that extra annotations such as spatiotemporal information~\cite{porrello2020robust} and part labels~\cite{he2019part, meng2020parsing}, can improve the mAP, but they are not robust in the cross-camera generalization. Compared with the above methods, our method reaches a high performance on both mAP and mCGM.

\begin{table}
\centering{
\caption{Analyses of cross-level stages on VeRi-776.}
\label{table:stage number}
\renewcommand{\arraystretch}{0.9}
\begin{tabular}{c|c|c|c|c|c}
\toprule
Stage2&Stage3&Stage4&mCGM&mAP&CMC@5\\
\midrule
&&\checkmark&0.620&0.819&0.985\\
&\checkmark&\checkmark&0.623&0.825&0.986\\
\checkmark&\checkmark&\checkmark&\textbf{0.630}&\textbf{0.831}&\textbf{0.989}\\
\bottomrule
\end{tabular}
}
\end{table}

\begin{table}
\centering{
\caption{Analyses of regional partition methods on VeRi-776.}
\label{table:partition method}
\renewcommand{\arraystretch}{0.9}
\begin{tabular}{l|c|c|c}
\toprule
Pooling Partition&mCGM&mAP&CMC@5\\
\midrule
Grid &0.611&0.817&0.985\\
Vertical&0.614&0.818&0.985\\
Horizontal&0.622&0.822&0.985\\
Progressive Center&\textbf{0.630}&\textbf{0.831}&\textbf{0.989}\\
\bottomrule
\end{tabular}
}
\end{table}

\subsection{Performance Analysis}
\textbf{Analysis of cross-level stages.} We conduct experiments on the effectiveness of complementary features in different lower-level stages. In these experiments, we gradually add the features from the lower-level stage as complements for those in the higher-level stage. In~\tabref{table:stage number}, it is obvious that the performances in all measures gain an improvement as the number of stages increases. It is shown that high-level ones can take meaningful complementary semantic information from the lower level features when adopting the stage-by-stage aggregation strategy.

\textbf{Analysis of regional partition methods.} As shown in~\tabref{table:partition method}, we exhibit four partition methods in experiments, which are grid partition, vertical partition, horizontal partition, and our progressive center pooling partition. Compared with three conventional partitions, our proposed method achieves a notable performance improvement on three measures due to its better ability of region alignment.

\textbf{Analysis of graph-based relation module.} To verify the effectiveness of GRM, we replace it with the simple concatenation operation (w/o GRM) on VeRi-776. As shown in~\tabref{table:performance on VeRi}, that without GRM is conspicuously lower than with GRM. To explain why heterogeneous features are fused by the GRM in a bottom-top manner, we design a new model which fuses futures into a unified graph for final output. This new model achieves 0.803 mAP in the VeRi-776 dataset, which shows a clear drop compared to our bottom-top model of 0.831 mAP.

\textbf{Analysis of model parameters and visualization.} Details are shown in the supplementary material.

\section{Conclusion}
In this paper, we propose a novel Heterogeneous Relation Complement Network (HRCN) to take cross-level features and region-specific features as complements for high-level features. With these heterogeneous complementary features, the final representation can focus on more discriminative regions which are critical to recognize the identity in diverse viewpoints. Experimental results reveal that our proposed approach reaches new state-of-the-arts on VeRi-776, VehicleID, and VERI-Wild benchmarks. Moreover, we propose a novel and efficient measure named cross-camera generalization measure (CGM) to evaluate cross-camera generalization capability construct a benchmark of 9 state-of-the-art methods and our approach.

\section*{Acknowledgments}
This work was supported by grants from National Natural Science Foundation of China (61922006 and 61825101).

{\small
\bibliographystyle{ieee_fullname}
\bibliography{egbib}
}

\end{document}